\documentclass[10pt, a4paper, dvipsnames]{article}

\usepackage[final]{lrec2026} 

\title{Integrating Arithmetic Learning Improves\\Mathematical Reasoning in Smaller Models}

\name{Neeraj Gangwar, Suma Bhat, Nickvash Kani} 

\address{Electrical and Computer Engineering \\
         University of Illinois Urbana-Champaign, IL, USA \\
         \texttt{\{gangwar2,spbhat2,kani\}@illinois.edu}\\}

\usepackage{hyperref}
\usepackage{url}
\usepackage{booktabs}
\usepackage[T1]{fontenc}
\usepackage{microtype}
\usepackage{xcolor}
\usepackage{graphicx}
\usepackage{svg}
\usepackage{multirow}
\usepackage{multicol}
\usepackage{amssymb}
\usepackage{pifont}
\usepackage{subfig}
\usepackage{todonotes}
\usepackage{breakurl}
\usepackage{listings}
\usepackage{fancyvrb}
\usepackage{textgreek}
\usepackage{wrapfig}
\usepackage{titlesec}
\usepackage{tikz}
\usetikzlibrary{calc}

\newcommand{\tikzxmark}{%
\tikz[scale=0.18] {
    \draw[line width=1.2,line cap=round] (0,0) to [bend left=6] (1,1);
    \draw[line width=1.2,line cap=round] (0.2,0.95) to [bend right=3] (0.8,0.05);
}}
\newcommand{\tikzcmark}{%
\tikz[scale=0.18] {
    \draw[line width=1.2, line cap=round] (0.25, 0) to [bend left=10] (1,1);
    \draw[line width=1.4, line cap=round] (0, 0.35) to [bend right=1] (0.23,0);
}}

\newcommand{\cmark}{\tikzcmark}
\newcommand{\xmark}{\tikzxmark}
\newcommand\poscolor{OliveGreen}
\newcommand\negcolor{Red}

\newcommand{\gsm}{GSM8k}
\newcommand{\gsmo}{\gsm{} (Orig.)}
\newcommand{\gsmd}{\gsm{} (Dist.)}

\newcommand\tulu{\textsc{T\"{u}lu 3 SFT}}
\newcommand\tulun{\textsc{T\"{u}lu 3}}

\usepackage{amsmath,amsfonts,bm}

\newcommand{\figleft}{{\em (Left)}}

\newcommand{\figright}{{\em (Right)}}
\newcommand{\figtop}{{\em (Top)}}
\newcommand{\figbottom}{{\em (Bottom)}}

\def\eqref#1{equation~\ref{#1}}

\def\1{\bm{1}}

\DeclareMathAlphabet{\mathsfit}{\encodingdefault}{\sfdefault}{m}{sl}
\SetMathAlphabet{\mathsfit}{bold}{\encodingdefault}{\sfdefault}{bx}{n}

\abstract{
    While large models pre-trained on high-quality data exhibit excellent performance on mathematical reasoning (e.g., GSM8k, MultiArith), it remains challenging to specialize smaller models for these tasks. Common approaches to address this challenge include knowledge distillation from large teacher models and data augmentation (e.g., rephrasing questions and generating synthetic solutions). Despite these efforts, smaller models struggle with arithmetic computations, leading to errors in mathematical reasoning. In this work, we leverage a synthetic arithmetic dataset generated programmatically to enhance the reasoning capabilities of smaller models. We investigate two key approaches to incorporate this dataset: (1) intermediate fine-tuning, in which a model is fine-tuned on the arithmetic dataset before training it on a reasoning dataset, and (2) integrating the arithmetic dataset into an instruction-tuning mixture, allowing the model to learn arithmetic skills alongside general instruction-following abilities. Our experiments on multiple reasoning benchmarks demonstrate that incorporating an arithmetic dataset, whether through targeted fine-tuning or within an instruction-tuning mixture, enhances models' arithmetic capabilities, thereby improving their mathematical reasoning performance.\\
    \newline
}

\begin{document}

\maketitleabstract

\section{Introduction}
Scaling the model and data sizes has had a tremendous effect on performance across various natural language processing (NLP) tasks \citep[e.g.,][]{grattafiori2024llama, liu2024deepseek, team2024gemma, team2024qwen2, Abdin2024Phi3TR, lambert2024t, jaech2024openai, chowdhery2023palm, achiam2023gpt, touvron2023llama2, jiang2023mistral}. These pre-trained models can learn from a few demonstrations using in-context learning and do not require task-specific fine-tuning \citep{brown2020language}. They also benefit from generating a sequence of reasoning steps before arriving at the final answer. These strategies have been particularly effective for mathematical reasoning \citep[][]{wei2022chain, nye2022show, fu2022complexity, zhou2022least}. While large models exhibit excellent performance on various reasoning benchmarks, smaller models remain essential due to their efficiency and adaptability. They require significantly fewer computational resources, making them ideal for deployment with limited infrastructure. They also enable faster inference and lower latency, which is crucial for real-time applications. Additionally, they can be fine-tuned more efficiently for specific tasks without the high costs associated with training massive models.

\begin{figure}
    \centering
    \scalebox{0.85}{
    \begin{tikzpicture}[
    font=\fontfamily{phv}\selectfont\footnotesize,
    text=darkgray!90,
]
    \node [draw=CadetBlue!15, fill=CadetBlue!15, rounded corners, text width=0.99\linewidth] (QUESTION) {
    Dylan needed chicken sausages and fish sausages to make sausage buns at a party. He bought 38 chicken sausages and 6 more fish sausages than chicken sausages. How many sausages did Dylan buy in all?
    };

    \node [anchor=north west] at ($(QUESTION.north west) + (0, 15pt)$) {\bfseries Question:};

    \node [draw=Red!10, fill=Red!10, rounded corners, text width=0.99\linewidth, below=of QUESTION, yshift=7pt] (DFT) {
    Dylan bought 38 chicken sausages.\\He bought 6 more fish sausages than chicken sausages.\\So he bought 6 + 38 = 42 fish sausages.\\In all, he bought 38 + 42 = 74 sausages.\\The answer is 74. \hfill \textcolor{Red}{\scriptsize \xmark}
    };

    \node [anchor=north west] at ($(DFT.north west) + (0, 15pt)$) {\bfseries No arithmetic dataset during training:};

    \node [draw=YellowGreen!15, fill=YellowGreen!15, rounded corners, text width=0.99\linewidth, below=of DFT, yshift=7pt] (IFT) {
    Dylan bought 38 chicken sausages.\\He bought 6 more fish sausages than chicken sausages.\\So he bought 6 + 38 = 44 fish sausages.\\In total, he bought 38 + 44 = 82 sausages.\\The answer is 82. \hfill \textcolor{OliveGreen}{\cmark}
    };

    \node [anchor=north west] at ($(IFT.north west) + (0, 15pt)$) {\bfseries Including arithmetic dataset during training:};
\end{tikzpicture}
    }
    \caption{An example from the \gsm{} test set and its solutions generated by \figtop{} FlanT5-Large directly fine-tuned on \gsm{}, and \figbottom{} FlanT5-Large fine-tuned on an arithmetic dataset before training it on \gsm{}.}
    \label{fig:arith_error_example}
\end{figure}

Traditionally, smaller pre-trained models are adapted for downstream tasks through supervised fine-tuning on a dataset or a combination of datasets formatted as instructions \citep[also referred to as instruction tuning;][]{wei2022finetuned}. While this approach is effective for simpler tasks, it falls short when applied to mathematical reasoning, as smaller models struggle to achieve strong performance. This challenge arises because math reasoning datasets, such as  GSM8k \citep{cobbe2021training}, consist of a small number of reasoning problems, typically paired with a solution. The scarcity of training examples makes it difficult for the model to capture the complexity of mathematical reasoning. To overcome this issue, a widely explored research direction is to distill knowledge from large pre-trained teacher models into smaller student models. Some methods use questions from existing training datasets and use prompting to generate multiple solutions per question for fine-tuning smaller models \citep{ho2023large, magister2023teaching, fu2023specializing, hsieh2023distilling, yuemammoth}. Others use various techniques such as rephrasing questions to create more examples \citep{yu2024metamath, huang2024key, tong2024dart, kaur2024instruct} or generating multiple views of a solution \citep{liang2024mint}, for better reasoning performance.

Although these methods boost mathematical reasoning performance in smaller models, the models may still struggle with arithmetic computations. In many cases, models generate the correct reasoning steps but arrive at an incorrect final answer due to numerical computation errors. See Figure~\ref{fig:arith_error_example} for an example. Some approaches address this issue by integrating external tools \citep{cobbe2021training, schick2023toolformer} or utilizing programs \citep{pmlr-v202-gao23f, chen2023program, ye_satlm_2023} to delegate these computations to external systems such as calculators. In this work, we explore whether the model performance can be improved by directly mitigating these errors without relying on external tools. Previous research has investigated ways to enhance arithmetic skills in Transformer-based models \citep{mcleish2024transformers, liu2023goat, yang2023gpt}. But the effective transfer of these skills to downstream tasks, such as mathematical reasoning, remains largely unexplored. To bridge this gap, we examine how improving arithmetic skills strengthens a model's mathematical reasoning.

In this work, we use a synthetic arithmetic dataset generated programmatically to enhance mathematical reasoning in small-frame models. We investigate two approaches for incorporating this dataset: (1) intermediate fine-tuning, in which a model is fine-tuned on the arithmetic dataset before training it on a reasoning dataset, and (2) integrating the arithmetic dataset into an instruction-tuning mixture. The first approach is inspired by transfer learning, as prior research shows that fine-tuning a model on a related dataset before training it on the target task can significantly improve its performance \citep{phang2018sentence, vu-etal-2020-exploring, pruksachatkun2020intermediate}. The second approach aligns with post-training techniques to refine pre-trained models by exposing them to diverse tasks, helping them acquire new skills and adapt better to various downstream tasks \citep{wei2021finetuned, chung2024scaling, lambert2024t}.

Empirical observations using several mathematical datasets lead to the following key takeaways:
\begin{itemize}
    \item Models fine-tuned on an arithmetic dataset before a reasoning dataset perform better than those fine-tuned directly on the reasoning dataset. Additionally, arithmetic datasets can be generated programmatically, eliminating the need for manual resources.

    \item Based on our observations with multiple datasets with varying mathematical reasoning tasks, we find that intermediate fine-tuning results in better out-of-domain generalization.

    \item Including an arithmetic dataset into the instruction-tuning mixture leads to better few-shot performance on multiple mathematical reasoning benchmarks.

    \item These models exhibit better robustness to numerical variations such as numerical substitution and digit expansion than models instruction-tuned on a mixture without the arithmetic dataset.
\end{itemize}

This work highlights the importance of explicit arithmetic training as a key factor in improving mathematical reasoning in smaller models. Our source code and datasets are publicly available.\footnote{\url{https://github.com/neerajgangwar/reasoning-with-arith}}

\section{Related Work}
\paragraph{Model~Specialization~via~Distillation.}
Adapting a pre-trained model for a downstream task has traditionally been done through task-specific fine-tuning. However, this approach does not work for mathematical reasoning tasks because datasets such as GSM8k do not contain enough examples to capture the complexity of mathematical reasoning. Several works have focused on distilling multi-step reasoning solutions from large language models (LLMs) to overcome this limitation. \citet{fu2023specializing} prompt Codex \citep{chen2021evaluating} to generate multiple multi-step solutions for examples in the GSM8k training set and fine-tune FlanT5 on the ones that lead to the correct final answer. \citet{hsieh2023distilling} use PaLM-540B \citep{chowdhery2023palm} for generating solutions and fine-tune T5 \citep{raffel2020exploring} in a multi-task setting to generate the labels and rationale. \citet{liu2023tinygsm} use GPT-3.5-turbo to generate synthetic GSM8k-like examples. \citet{yuemammoth} show that a hybrid of chain-of-thought and program-of-thought solutions performs better than using either format individually. In addition to using LLMs to generate more solutions, \citet{yu2024metamath} use LLM rephrasing and backward reasoning to augment questions and create a new dataset, MetaMathQA. Several studies \citep{huang2024key, tong2024dart, kaur2024instruct} have built on these techniques by using larger teacher models to create more efficient reasoning datasets. In contrast, our approach focuses on a dataset that can be generated programmatically without using external models and addresses a specific weakness in the model.

\vspace{-0.5em}
\paragraph{Transfer Learning.}
Transfer learning has played a pivotal role in NLP. \citet{vu-etal-2020-exploring} and \citet{pruksachatkun2020intermediate} study the effect of intermediate fine-tuning on the model's performance on a target task. \citet{conneau2019cross} explore cross-lingual model pre-training and show improvements in natural language inference and machine translation. \citet{razdaibiedina2023progressive} introduce progressive prompts, a continual learning approach with forward transfer without catastrophic forgetting. Training on large multi-task mixtures is also a common trend in NLP \citep{aribandi2022ext5, wei2022finetuned, chung2024scaling, lambert2024t}. Another research direction explores identifying relevant examples for a given downstream task from a huge collection of datasets, like P3 \citep{sanh2021multitask}. These methods create embeddings for all examples of interest using hidden states \citep{ivison2023data} or gradients \citep{xia2024less}.

\section{Methodology}
In this study, we investigate two approaches for transferring arithmetic skills to the more complex domain of mathematical reasoning: (1) intermediate fine-tuning on an arithmetic dataset, and (2) incorporating the arithmetic dataset during instruction tuning.

\vspace{-0.5em}
\paragraph{Intermediate Fine-Tuning.}
Fine-tuning a model on an intermediate task before a downstream task can improve the model's performance on the latter \citep{phang2018sentence, vu-etal-2020-exploring, pruksachatkun2020intermediate}. The downstream task is also referred to as the target task. This is called intermediate fine-tuning and can lead to successful knowledge transfer for similar intermediate and target tasks. Building on prior work on transfer learning, we use intermediate fine-tuning to transfer arithmetic abilities to mathematical reasoning.

Unlike natural language, mathematical computations are precise and do not contain redundancies, typically requiring more training examples for effective learning than NLP tasks. When a model is fine-tuned on a reasoning dataset, it must simultaneously learn to generate correct reasoning steps and arithmetic computations. Moreover, datasets such as GSM8k contain a limited number of examples, restricting both the quantity and diversity of arithmetic computations. This limitation often leads to arithmetic errors during inference. To address these challenges, we adopt a two-step training approach. First, we fine-tune the model on an arithmetic dataset, allowing it to learn a broad range of numerical computations across diverse values. This intermediate fine-tuning ensures that the model develops a strong foundation in arithmetic. Following this, the model is fine-tuned on a reasoning dataset, where it focuses on applying its pre-learned arithmetic skills rather than learning them from limited examples. This two-step process yields fewer arithmetic errors during mathematical reasoning than models directly fine-tuned on a reasoning dataset, leading to more accurate and reliable performance.

\vspace{-0.5em}
\paragraph{Instruction Tuning.}
We also explore the impact of incorporating an arithmetic dataset during post-training, which involves additional training on vast corpora of text. In particular, we focus on instruction tuning, a post-training technique to enhance the instruction-following ability of pre-trained models \citep{wei2021finetuned, chung2024scaling}. This process involves fine-tuning a pre-trained model on a diverse set of instructions and the corresponding responses. The fine-tuning mixture typically contains examples from a wide range of tasks, including the mathematical reasoning domain. To strengthen the arithmetic capabilities, we include a synthetic arithmetic dataset in this mixture. Similar to intermediate fine-tuning, this process improves the model's arithmetic proficiency and enhances mathematical reasoning performance by enabling more accurate numerical computations within reasoning tasks.

\section{Datasets}
\subsection{Arithmetic Dataset}
We refer to \citet{liu2023goat} for generating our dataset programmatically. The dataset from \citet{liu2023goat} contains basic arithmetic operations: addition, subtraction, multiplication, and division. We extend it to include problems on fractions and percentages. The datasets used in this work do not require computations over large numbers; hence, using the GSM8k training set as a representative benchmark, we limit operand lengths to seven digits. Furthermore, we use log-uniform sampling to ensure that the dataset is not skewed towards numbers with more digits. This dataset contains nearly 1.3M examples. See Appendix~\ref{app:synthetic_arithmetic_dataset} for samples from the generated dataset.

\subsection{GSM8k Training Dataset}
For the intermediate fine-tuning experiments, we use \gsm{} \citep{cobbe2021training} for model specialization. As the dataset does not have a validation set, we randomly sample 512 examples from the training set to create one. We use two versions of \gsm{} in this work.

\vspace{-0.5em}
\paragraph{Original.}
In the first version, we use the remaining examples from the training set. This dataset contains 6961 examples. We refer to this dataset as \gsmo{}.

\vspace{-0.5em}
\paragraph{Distilled.}
We generate a distilled dataset using the questions from \gsmo{} to evaluate if intermediate fine-tuning benefits tasks with large training datasets. This dataset is generated by prompting Mistral-7B \citep{jiang2023mistral} using the prompt from \citet{wei2022chain}. We generate 64 solutions per question and retain only those that lead to the correct final answer. Correctness is determined by matching the generated final answer with the ground truth. We do not verify reasoning chains, as manual verification is prohibitively resource-intensive. After removing duplicate solutions, the dataset contains approximately 175K examples. We refer to this dataset as \gsmd{}.

\subsection{Instruction Tuning Dataset}
Following the recent work of \citet{lambert2024t}, we use the \tulu{} mixture as the instruction tuning dataset for our work. It encompasses a broad spectrum of tasks, including general-purpose chat, complex mathematical reasoning, programming, instruction following, and persona-based synthetic data. This dataset contains nearly 1M examples, making it significantly smaller than the Flan collection \citep{longpre2023flan}. However, it contains more mathematical reasoning examples than datasets such as Flan-mini \citep{ghosal2023flacuna}.

\section{Intermediate Fine-Tuning}
\label{sec:intermediate_fine_tuning}

\subsection{Experimental Setup}
\paragraph{Models and Training Details.}
We experiment with both encoder-decoder and decoder-only architectures. We use FlanT5 \citep[base and large;][]{chung2024scaling} and GPT2 \citep[base, medium, and large;][]{radford2019language}. We first fine-tune these models on the arithmetic dataset for two epochs. To adapt these models for reasoning, we continue the training from these checkpoints on \gsm{}. The models are fine-tuned for 20 and 100 epochs on \gsmd{} and \gsmo{}, respectively. See Appendix~\ref{app:training_details_ift} for further details.

\vspace{-0.5em}

\paragraph{Baseline.}
For the baseline, we consider models that are directly fine-tuned on \gsm{} without any intermediate fine-tuning.

\vspace{-0.5em}
\paragraph{Evaluation Tasks.}
We evaluate our approach on the GSM8k test set. We also test out-of-domain generalization using the following datasets: (1) MultiArith \citep{roy-roth-2015-solving}, which contains problems focused on basic arithmetic operations and is relatively simpler than \gsm{}, (2) ASDiv \citep{miao-etal-2020-diverse} with diverse math problems focused on language usage patterns, and (3) SVAMP \citep{patel-etal-2021-nlp} with varying structures to ensure that a model cannot solve the problems by applying simple heuristics and ignoring question text.

\vspace{-0.5em}
\paragraph{Decoding.}
We use greedy decoding and self-consistency decoding during inference. For self-consistency decoding, nucleus sampling \citep{holtzman2019curious} is used with $T = 0.6$ and $p = 0.9$ to sample eight responses, and the most consistent final answer is chosen. As nucleus sampling is a stochastic decoding method, we repeat the evaluation three times and report the mean accuracy.

\subsection{Results}
\paragraph{In-Domain Performance.}
We first evaluate the models on the \gsm{} test set. Table~\ref{tab:main_results} shows the accuracy (\%) achieved by various models. These results indicate that FlanT5 benefits from intermediate fine-tuning, significantly improving the \gsm{} performance. Additionally, these results demonstrate that intermediate fine-tuning helps with both \gsmo{}, which has a small training set, and \gsmd{}, which already contains significantly more training examples.

\begin{table*}[t]
    \centering
    \setlength{\tabcolsep}{3pt}
    \resizebox{\textwidth}{!}{
    \begin{tabular}{p{1.5cm}lc|r@{ }lr@{ }lr@{ }lr@{ }l|r@{ }lr@{ }lr@{ }lr@{ }l}
    \toprule
    \multirow{3}{1.5cm}{\textbf{Training Dataset}} & \multirow{3}{*}{\textbf{Model}} & \multirow{3}{*}{\textbf{IFT}} & \multicolumn{8}{c}{\textbf{Greedy Decoding}} & \multicolumn{8}{|c}{\textbf{Self-Consistency Decoding}} \\
    {} & {} & {} & \multicolumn{2}{c}{\textbf{\gsm{}}} & \multicolumn{2}{c}{\textbf{MultiArith}} & \multicolumn{2}{c}{\textbf{ASDiv}} & \multicolumn{2}{c}{\textbf{SVAMP}} & \multicolumn{2}{|c}{\textbf{\gsm{}}} & \multicolumn{2}{c}{\textbf{MultiArith}} & \multicolumn{2}{c}{\textbf{ASDiv}} & \multicolumn{2}{c}{\textbf{SVAMP}} \\
    {} & {} & {} & \multicolumn{8}{c}{Accuracy ($\Delta$)} & \multicolumn{8}{|c}{Accuracy ($\Delta$)} \\
    \midrule
    \multirow{10}{1.5cm}{\gsmo{}} & \multirow{2}{*}{FlanT5-Base} & \xmark & \textcolor{white}{0}7.7 & {} & 17.2 & {} & \textcolor{white}{0}8.5 & {} & \textcolor{white}{0}6.6 & {} & \textcolor{white}{0}9.1 & {} & 17.4 & {} & \textcolor{white}{0}8.6 & {} & \textcolor{white}{0}7.7 & {} \\
    
    {} & {} & \cmark & 10.5 & {\footnotesize (\textcolor{\poscolor}{+2.8})} & 25.6 & {\footnotesize (\textcolor{\poscolor}{+8.4})} & 12.5 & {\footnotesize (\textcolor{\poscolor}{+4.0})} & 10.2 & {\footnotesize (\textcolor{\poscolor}{+3.6})} & 12.0 & {\footnotesize (\textcolor{\poscolor}{+2.9})} & 30.2 & {\footnotesize (\textcolor{\poscolor}{+12.8})} & 14.1 & {\footnotesize (\textcolor{\poscolor}{+5.5})} & 11.4 & {\footnotesize (\textcolor{\poscolor}{+3.7})} \\
    
    \cmidrule{2-19}
    
    {} & \multirow{2}{*}{FlanT5-Large} & \xmark & 12.9 & {} & 28.9 & {} & 15.7 & {} & 12.1 & {} & 14.7 & {} & 29.1 & {} & 16.8 & {} & 12.6 & {} \\
    
    {} & {} & \cmark & 17.1 & {\footnotesize (\textcolor{\poscolor}{+4.2})} & 45.0 & {\footnotesize (\textcolor{\poscolor}{+16.1})} & 20.8 & {\footnotesize (\textcolor{\poscolor}{+5.1})} & 16.2 & {\footnotesize (\textcolor{\poscolor}{+4.1})} & 18.6 & {\footnotesize (\textcolor{\poscolor}{+3.9})} & 47.2 & {\footnotesize (\textcolor{\poscolor}{+18.1})} & 21.4 & {\footnotesize (\textcolor{\poscolor}{+4.6})} & 16.6 & {\footnotesize (\textcolor{\poscolor}{+4.0})} \\
    
    \cmidrule{2-19}
    
    {} & \multirow{2}{*}{GPT2} & \xmark & \textcolor{white}{0}5.5 & {} & 12.8 & {} & \textcolor{white}{0}4.6 & {} & \textcolor{white}{0}3.9 & {} & \textcolor{white}{0}6.9 & {} & 18.3 & {} & \textcolor{white}{0}5.2 & {} & \textcolor{white}{0}4.7 & {} \\
    
    {} & {} & \cmark & \textcolor{white}{0}6.6 & {\footnotesize (\textcolor{\poscolor}{+1.1})} & 17.2 & {\footnotesize (\textcolor{\poscolor}{+4.4})} & \textcolor{white}{0}6.9 & {\footnotesize (\textcolor{\poscolor}{+2.3})} & \textcolor{white}{0}5.8 & {\footnotesize (\textcolor{\poscolor}{+1.9})} & \textcolor{white}{0}7.7 & {\footnotesize (\textcolor{\poscolor}{+0.8})} & 22.4 & {\footnotesize (\textcolor{\poscolor}{+4.1})} & \textcolor{white}{0}8.4 & {\footnotesize (\textcolor{\poscolor}{+3.2})} & \textcolor{white}{0}6.5 & {\footnotesize (\textcolor{\poscolor}{+1.8})} \\
    
    \cmidrule{2-19}
    
    {} & \multirow{2}{*}{GPT2-Medium} & \xmark & \textcolor{white}{0}7.9 & {} & 20.0 & {} & \textcolor{white}{0}8.8 & {} & \textcolor{white}{0}5.9 & {} & \textcolor{white}{0}9.0 & {} & 23.7 & {} & \textcolor{white}{0}9.3 & {} & \textcolor{white}{0}6.5 & {} \\
    
    {} & {} & \cmark & \textcolor{white}{0}7.4 & {\footnotesize (\textcolor{\negcolor}{-0.5})} & 26.7 & {\footnotesize (\textcolor{\poscolor}{+6.7})} & 10.1 & {\footnotesize (\textcolor{\poscolor}{+1.3})} & \textcolor{white}{0}6.1 & {\footnotesize (\textcolor{\poscolor}{+0.2})} & \textcolor{white}{0}7.7 & {\footnotesize (\textcolor{\negcolor}{-1.3})} & 32.8 & {\footnotesize (\textcolor{\poscolor}{+9.1})} & 11.2 & {\footnotesize (\textcolor{\poscolor}{+1.9})} & \textcolor{white}{0}7.1 & {\footnotesize (\textcolor{\poscolor}{+0.6})} \\
    
    \cmidrule{2-19}
    
    {} & \multirow{2}{*}{GPT2-Large} & \xmark & \textcolor{white}{0}8.5 & {} & 23.3 & {} & 11.0 & {} & \textcolor{white}{0}9.4 & {} & \textcolor{white}{0}9.1 & {} & 28.9 & {} & 11.9 & {} & 10.0 & {} \\
    
    {} & {} & \cmark & \textcolor{white}{0}7.1 & {\footnotesize (\textcolor{\negcolor}{-1.4})} & 26.7 & {\footnotesize (\textcolor{\poscolor}{+3.4})} & \textcolor{white}{0}8.3 & {\footnotesize (\textcolor{\negcolor}{-2.7})} & \textcolor{white}{0}7.7 & {\footnotesize (\textcolor{\negcolor}{-1.7})} & \textcolor{white}{0}8.8 & {\footnotesize (\textcolor{\negcolor}{-0.3})} & 25.2 & {\footnotesize (\textcolor{\negcolor}{-3.7})} & \textcolor{white}{0}9.4 & {\footnotesize (\textcolor{\negcolor}{-2.5})} & \textcolor{white}{0}8.9 & {\footnotesize (\textcolor{\negcolor}{-1.1})} \\
    
    \midrule
    
    \multirow{10}{1.5cm}{\gsmd{}} & \multirow{2}{*}{FlanT5-Base} & \xmark & 17.5 & {} & 31.1 & {} & 23.6 & {} & 20.4 & {} & 19.9 & {} & 33.9 & {} & 24.6 & {} & 20.2 & {} \\
    
    {} & {} & \cmark & 21.4 & {\footnotesize (\textcolor{\poscolor}{+3.9})} & 42.2 & {\footnotesize (\textcolor{\poscolor}{+11.1})} & 29.6 & {\footnotesize (\textcolor{\poscolor}{+6.0})} & 22.5 & {\footnotesize (\textcolor{\poscolor}{+2.1})} & 23.6 & {\footnotesize (\textcolor{\poscolor}{+3.7})} & 45.6 & {\footnotesize (\textcolor{\poscolor}{+11.7})} & 31.2 & {\footnotesize (\textcolor{\poscolor}{+6.6})} & 25.4 & {\footnotesize (\textcolor{\poscolor}{+5.2})} \\
    
    \cmidrule{2-19}
    
    {} & \multirow{2}{*}{FlanT5-Large} & \xmark & 22.4 & {} & 45.0 & {} & 29.1 & {} & 23.2 & {} & 24.9 & {} & 43.7 & {} & 30.2 & {} & 25.0 & {} \\
    
    {} & {} & \cmark & 27.7 & {\footnotesize (\textcolor{\poscolor}{+5.3})} & 73.3 & {\footnotesize (\textcolor{\poscolor}{+28.3})} & 40.0 & {\footnotesize (\textcolor{\poscolor}{+10.9})} & 33.3 & {\footnotesize (\textcolor{\poscolor}{+10.1})} & 30.1 & {\footnotesize (\textcolor{\poscolor}{+5.2})} & 76.1 & {\footnotesize (\textcolor{\poscolor}{+32.4})} & 41.4 & {\footnotesize (\textcolor{\poscolor}{+11.2})} & 34.9 & {\footnotesize (\textcolor{\poscolor}{+9.9})} \\
    
    \cmidrule{2-19}
    
    {} & \multirow{2}{*}{GPT2} & \xmark & 14.8 & {} & 44.4 & {} & 24.2 & {} & 17.9 & {} & 18.8 & {} & 46.9 & {} & 25.5 & {} & 18.7 & {} \\
    
    {} & {} & \cmark & 19.6 & {\footnotesize (\textcolor{\poscolor}{+4.8})} & 70.6 & {\footnotesize (\textcolor{\poscolor}{+26.2})} & 32.4 & {\footnotesize (\textcolor{\poscolor}{+8.2})} & 22.5 & {\footnotesize (\textcolor{\poscolor}{+4.6})} & 22.2 & {\footnotesize (\textcolor{\poscolor}{+3.4})} & 77.4 & {\footnotesize (\textcolor{\poscolor}{+30.5})} & 35.3 & {\footnotesize (\textcolor{\poscolor}{+9.8})} & 26.2 & {\footnotesize (\textcolor{\poscolor}{+7.5})} \\
    
    \cmidrule{2-19}
    
    {} & \multirow{2}{*}{GPT2-Medium} & \xmark & 21.4 & {} & 55.0 & {} & 27.7 & {} & 18.8 & {} & 24.8 & {} & 56.1 & {} & 29.8 & {} & 23.0 & {} \\
    
    {} & {} & \cmark & 25.9 & {\footnotesize (\textcolor{\poscolor}{+4.5})} & 72.8 & {\footnotesize (\textcolor{\poscolor}{+17.8})} & 37.7 & {\footnotesize (\textcolor{\poscolor}{+10.0})} & 26.5 & {\footnotesize (\textcolor{\poscolor}{+7.7})} & 29.6 & {\footnotesize (\textcolor{\poscolor}{+4.8})} & 80.0 & {\footnotesize (\textcolor{\poscolor}{+23.9})} & 40.6 & {\footnotesize (\textcolor{\poscolor}{+10.8})} & 30.9 & {\footnotesize (\textcolor{\poscolor}{+7.9})} \\
    
    \cmidrule{2-19}
    
    {} & \multirow{2}{*}{GPT2-Large} & \xmark & 19.9 & {} & 52.8 & {} & 29.0 & {} & 22.3 & {} & 24.1 & {} & 56.5 & {} & 31.1 & {} & 23.5 & {} \\
    
    {} & {} & \cmark & 24.3 & {\footnotesize (\textcolor{\poscolor}{+4.4})} & 75.0 & {\footnotesize (\textcolor{\poscolor}{+22.2})} & 37.1 & {\footnotesize (\textcolor{\poscolor}{+8.1})} & 30.2 & {\footnotesize (\textcolor{\poscolor}{+7.9})} & 28.8 & {\footnotesize (\textcolor{\poscolor}{+4.7})} & 81.7 & {\footnotesize (\textcolor{\poscolor}{+25.2})} & 41.1 & {\footnotesize (\textcolor{\poscolor}{+10.0})} & 33.7 & {\footnotesize (\textcolor{\poscolor}{+10.2})} \\
    
    \bottomrule
    \end{tabular}
    }

    \caption{Accuracy (\%) of models fine-tuned on the \gsm{} datasets, with and without prior intermediate fine-tuning (IFT) on the arithmetic dataset. We report results for both greedy decoding and self-consistency decoding. Performance on MultiArith, ASDiv, and SVAMP is included to demonstrate that IFT maintains out-of-domain generalization.}
    \label{tab:main_results}
\end{table*}

For GPT2, we observe a slight decline in reasoning performance when the model is specialized in reasoning using \gsmo{} after fine-tuning it on the arithmetic dataset. However, this issue does not arise with \gsmd{}, where we see performance gains comparable to those of FlanT5. We attribute this behavior to the fact that intermediate fine-tuning optimizes the model for arithmetic tasks, making it more challenging to adapt to other tasks compared to the original model. A larger fine-tuning dataset mitigates this issue, as demonstrated by the significant performance improvement when reasoning specialization is performed using \gsmd{}.

\vspace{-0.5em}
\paragraph{Out-of-Domain Performance.}
Next, the models fine-tuned on \gsm{} are evaluated on MultiArith, ASDiv, and SVAMP, and the results are shown in Table~\ref{tab:main_results}. These results indicate that intermediate fine-tuning does not hurt out-of-domain generalization. Conversely, the models fine-tuned on the arithmetic dataset first generalize better than those directly fine-tuned on \gsm{}.

\vspace{-0.5em}
\paragraph{Arithmetic in Reasoning Context.}
While the models fine-tuned on the arithmetic dataset excel at basic arithmetic tasks compared to their original versions, do these skills transfer to reasoning tasks when they are fine-tuned on reasoning datasets? Moreover, are they the reason behind the better reasoning performance, as shown in Table~\ref{tab:main_results}? Accuracy on the test sets from various reasoning datasets, such as GSM8k and MultiArith, does not directly capture this, as an incorrect final answer can stem from multiple factors beyond arithmetic errors. To better understand the impact of arithmetic computations, we specifically look at them within the reasoning process, ensuring that all other reasoning steps remain correct.

We use the \gsm{} test set and identify the arithmetic computations using the calculation annotations (enclosed within \texttt{<<>>}). Given a question and its solution up to an annotation, the models are prompted to generate the next tokens, which are then compared to the ground truth. We define the accuracy of these tokens as GSM8k arithmetic accuracy. See Appendix~\ref{app:gsm8k_arith_details} for more details. Figure~\ref{fig:gsm8k_arith_perf} shows how well the models from Table~\ref{tab:main_results} handle arithmetic computations within reasoning contexts. These results suggest that intermediate fine-tuning reduces arithmetic errors compared to the models fine-tuned directly on GSM8k, leading to an average improvement of 11.7\% in arithmetic computations. Even GPT2, when fine-tuned on \gsmo{}, makes fewer arithmetic errors with intermediate fine-tuning. These results support the hypothesis that the slight decline in the reasoning performance of GPT2 fine-tuned on \gsmo{} after the intermediate fine-tuning results from the model being optimized for arithmetic tasks, potentially making it challenging to adapt the model for reasoning tasks with a smaller training set. A larger training dataset, \gsmd{} in this case, mitigates the issue.

\begin{figure}[t]
    \centering
    \resizebox{0.95\linewidth}{!}{
        \Large\fontfamily{phv}\selectfont
        \includesvg[]{img/gsm8k_arith_perf.svg}
    }
    \caption{\gsm{} arithmetic accuracy or the ability of the fine-tuned models to generate correct arithmetic computations in reasoning contexts. This evaluation is performed on the \gsm{} test set.}
    \label{fig:gsm8k_arith_perf}
\end{figure}

\begin{figure*}
    \resizebox{0.93\textwidth}{!}{
        \Large\fontfamily{phv}\selectfont
        \includesvg[]{img/gsm8k_vs_ift_epochs.svg}
    }
    \caption{GSM8k arithmetic accuracy \figleft{} and GSM8k test accuracy \figright{} of the models fine-tuned on GSM8k after the intermediate fine-tuning for different number of epochs. The GSM8k arithmetic performance saturates after two epochs of intermediate fine-tuning and yields no further improvement in the GSM8k performance.}
    \label{fig:gsm8k_vs_ift_epochs}
\end{figure*}

\vspace{-0.5em}
\paragraph{Prolonged Intermediate Fine-Tuning.}
We next investigate the impact of extending the intermediate fine-tuning beyond two epochs. We find that fine-tuning models on the arithmetic dataset for additional epochs provides no benefit and even adversely affects the reasoning performance when models are later fine-tuned on a reasoning dataset. We attribute this outcome to two factors. First, prolonged fine-tuning on the arithmetic dataset further optimizes the model for arithmetic tasks, limiting its adaptability for other tasks. Second, the model learns the computations for solving GSM8k problems within the first few epochs, and further training does not improve this ability. Figure~\ref{fig:gsm8k_vs_ift_epochs} illustrates GSM8k arithmetic accuracy and overall performance as a function of intermediate fine-tuning epochs. The results show that extended intermediate fine-tuning does not improve GSM8k arithmetic accuracy and specializes models for arithmetic tasks. This leads to no improvement or decline in GSM8k performance.

\section{Instruction Tuning}
\label{sec:instruction_tuning}

\subsection{Experimental Setup}
\paragraph{Model and Training Details.}
As LLMs predominantly use the decoder-only architecture, we use GPT2 \citep[large;][]{radford2019language} for this experiment. The models are fine-tuned for five epochs. See Appendix~\ref{app:training_details_it} for training details.

\vspace{-0.5em}
\paragraph{Baselines.}
We use the following two baselines: (1) the pre-trained model and (2) the model fine-tuned only on the \tulu{} mixture.

\vspace{-0.5em}
\paragraph{Evaluation Tasks.}
We use nine math reasoning datasets to evaluate the impact of including an arithmetic dataset in the instruction tuning mixture. Similar to the intermediate fine-tuning experiments, we evaluate the models on GSM8k, ASDiv, SVAMP, and MultiArith. Additionally, we consider the following datasets from MAWPS \citep{koncel2016mawps}: (1) AddSub \citep{hosseini-etal-2014-learning}, which is a collection of addition and subtraction problems, (2) SingleEq \citep{roy2015reasoning}, which contains single equation problems, (3) SingleOp \citep{roy2015reasoning} with single operation arithmetic problems, and (4) SimulEq \citep{kushman2014learning} with multiple equation math word problems. AQuA \citep{ling2017program}, which contains algebraic problems in multiple-choice format, is also used for evaluation.

\begin{table*}
    \centering
    \begin{small}
    \begin{tabular}{ccrrrrrr}
    \toprule
    \textbf{\tulu{}} & \textbf{Arith.} & \textbf{GSM8k} & \textbf{ASDiv} & \textbf{SVAMP} & \textbf{MAWPS} & \textbf{AQuA} & \textbf{GSM-Plus} \\
    \midrule
    \multicolumn{8}{c}{\it Greedy Decoding} \\
    \midrule
    \xmark & \xmark & 1.5 & 2.3 & 3.3 & 2.7 & 16.5 & 2.0 \\
    \cmark & \xmark & 15.0 & 24.5 & 19.9 & 30.3 & 18.1 & 7.0 \\
    \cmark & \cmark & \textbf{16.3} & \textbf{36.9} & \textbf{28.1} & \textbf{43.3} & \textbf{22.8} & \textbf{9.2} \\
    \midrule
    \multicolumn{8}{c}{\it Self-Consistency Decoding} \\
    \midrule
    \xmark & \xmark & 2.5 & 2.6 & 4.1 & 2.4 & 18.0 & 2.0 \\
    \cmark & \xmark & 17.3 & 30.9 & 26.4 & 37.4 & \textbf{21.5} & 9.4 \\
    \cmark & \cmark & \textbf{19.6} & \textbf{42.6} & \textbf{33.9} & \textbf{50.1} & 19.8 & \textbf{11.4} \\
    \bottomrule
    \end{tabular}
    \end{small}
    \caption{Accuracy (\%) achieved by GPT2-Large when instruction-tuned on the \tulu{} mixture with and without including the synthetic arithmetic dataset. The first rows under both greedy decoding and self-consistency decoding denote the pre-trained model.}
    \label{tab:inst_tuning_results}
\end{table*}

We also evaluate the robustness of the models against various perturbations. For this purpose, we use two datasets: GSM-Plus \citep{li2024gsm} and GSM-Symbolic \citep{mirzadeh2024gsm}. GSM-Plus is an adversarial grade school math dataset that introduces five variations for each problem in the \gsm{} test set: numerical variation, arithmetic variation, rephrasing, distractor insertion, and omissions of necessary statements. GSM-Symbolic contains 100 test problems from GSM8k for which variations can be systematically generated by altering numerical values or proper names. As this work focuses on arithmetic computations, we generate 50 variations by modifying the numerical values in the original problems for our experiments.

\vspace{-0.5em}
\paragraph{Decoding.}
We use few-shot prompting to evaluate the models. Four exemplars are used in the prompts due to the maximum sequence length limit in GPT2. We use exemplars from the prompts used in \citet{wei2022chain}. See Appendix~\ref{app:math_prompt} for more information. We follow the same strategies for decoding as described in Section~\ref{sec:intermediate_fine_tuning}.

\subsection{Results}
Table~\ref{tab:inst_tuning_results} shows the results of instruction tuning GPT2-Large with and without including the synthetic arithmetic dataset in the fine-tuning mixture. The model fine-tuned on the \tulu{} mixture and synthetic arithmetic examples achieves better performance across math reasoning datasets, except AQuA, than the model fine-tuned only on the \tulu{} mixture, with both greedy decoding and self-consistency decoding. For self-consistency decoding, the former outperforms the latter in all three evaluation attempts across datasets. We also compute the GSM8k arithmetic accuracy for these models and find a 3\% increase in accuracy by including the arithmetic dataset in the fine-tuning mixture. See Appendix~\ref{app:mawps_detailed_results} for the results on individual datasets in MAWPS.

Neither model performs well on AQuA, and the performance of all models is close to random choice. Randomly choosing an option leads to an average accuracy of 19.9\% $\pm$ 2.7\% after 100 trials.

\begin{figure}[t]
    \centering
    \resizebox{\linewidth}{!}{
        \large\fontfamily{phv}\selectfont
        \includesvg[]{img/gsm_plus_category_wise_maj.svg}
    }
    \caption{Performance of the pre-trained and instruction-tuned GPT2-Large models on GSM-Plus for different perturbation types using self-consistency decoding. The model fine-tuned on \tulu{} mixture and the arithmetic dataset performs better across different perturbation types. The percentages above the bars represent the performance drop relative to the original GSM8k dataset, as shown in Table~\ref{tab:inst_tuning_results}.}
    \label{fig:gsm_plus_accuracy}
\end{figure}

\subsection{Robustness to Perturbations}
\paragraph{GSM-Plus.}
The overall accuracy is shown in Table~\ref{tab:inst_tuning_results}. The breakup of the overall performance by perturbation types is illustrated in Figure~\ref{fig:gsm_plus_accuracy}. The model fine-tuned on the \tulu{} and arithmetic mixture performs better than the model fine-tuned only on the \tulu{} mixture across different perturbations. We further investigate the performance drop for the two models relative to the original \gsm{} dataset. In particular, we are interested in numerical variations, which include numerical substitutions, digit expansions, and integer-decimal-fraction conversions. For these perturbations, we find that the model fine-tuned on the \tulu{} mixture and the arithmetic dataset sees a lower performance drop relative to the original GSM8k dataset than the model fine-tuned only on the \tulu{} mixture.

\begin{table}[t]
    \centering
    \begin{small}
    \begin{tabular}{ccrrr}
    \toprule
    \textbf{\tulun{}} & \multirow{2}{*}{\textbf{Arith.}} & \textbf{GSM8k} & \textbf{GSM-Symb.} & $\Delta$ \\
    \textbf{SFT} & {} & Acc. (\%) & Acc. (\%) & (\%) \\
    \midrule
    \cmark & \xmark & 15.3 & 8.5 $\pm$ 1.6 & $-$44.4 \\
    \cmark & \cmark & 22.0 & 16.3 $\pm$ 2.1 & \textbf{$\mathbf{-}$25.9} \\
    \bottomrule
    \end{tabular}
    \end{small}
    \caption{Performance of the instruction-tuned models using self-consistency decoding when evaluated on the original GSM8k problems vs the same problems in GSM-Symbolic. $\Delta$ indicates the performance drop.}
    \label{tab:gsm_symbolic_results}
\end{table}

\begin{table}[t]
    \centering
    \begin{small}
    \begin{tabular}{lrr}
    \toprule
    \textbf{Dataset} & \textbf{w/o Arith.} & \textbf{w/ Arith.} \\
    \midrule
    IFEval & {} & {} \\
    \hspace{1em}Strict (prompt level) & 0.3549 & 0.3420 \\
    \hspace{1em}Strict (instruction level) & 0.4964 & 0.4904 \\
    \hspace{1em}Loose (prompt level) & 0.3826 & 0.3826 \\
    \hspace{1em}Loose (instruction level) & 0.5252 & 0.5252 \\
    \midrule
    BigBench-Hard (Avg.) & 19.04\% & 18.03\% \\
    \bottomrule
    \end{tabular}
    \end{small}
    \caption{Performance of instruction-tuned GPT2-Large (from Section~\ref{sec:instruction_tuning}) on IFEval and BigBench-Hard. See Table~\ref{tab:bbh_detailed_results} in the appendix for the detailed performance on BigBench-Hard.}
    \label{tab:ifeval_bbh_results}
\end{table}
\begin{table*}[t]
    \centering
    \begin{small}
    \setlength{\tabcolsep}{3pt}
    \begin{tabular}{rrrrrr}
    \toprule
    \textbf{GSM8k} & \textbf{ASDiv} & \textbf{SVAMP} & \textbf{MAWPS} & \textbf{AQuA} & \textbf{GSM-Plus} \\
    \midrule
    12.6 & 36.3 & 29.7 & 47.1 & 22.1 & 7.4 \\
    \bottomrule
    \end{tabular}
    \end{small}
    \caption{Accuracy (\%) achieved by GPT2-Large (with greedy decoding) when instruction-tuned on a dataset composed of 90\% of the \tulu{} examples combined with the arithmetic dataset.}
    \label{tab:token_budget_results}
\end{table*}
\begin{table*}[t]
    \centering
    \begin{small}
    \begin{tabular}{l|cccc|cccc}
    \toprule
    \multirow{3}{*}{\textbf{Model}} & \multicolumn{4}{c|}{\textbf{Greedy Decoding}} & \multicolumn{4}{c}{\textbf{Self-Consistency Decoding}} \\
    {} & \textbf{\gsm{}} & \textbf{MultiArith} & \textbf{ASDiv} & \textbf{SVAMP} & \textbf{\gsm{}} & \textbf{MultiArith} & \textbf{ASDiv} & \textbf{SVAMP} \\
    {} & \multicolumn{4}{c|}{Accuracy} & \multicolumn{4}{c}{Accuracy} \\
    \midrule
    FlanT5-Base & 21.2 & 46.1 & 30.4 & 23.6 & 23.7 & 50.4 & 31.4 & 25.8 \\
    GPT2 & 19.4 & 67.8 & 31.2 & 20.9 & 23.1 & 75.4 & 34.3 & 25.6 \\
    \bottomrule
    \end{tabular}
    \end{small}
    \caption{Accuracy (\%) achieved by models fine-tuned on the GSM8k datasets after fine-tuning them on the decontaminated arithmetic dataset. Here, \textit{decontamination} refers to the removal of arithmetic computations appearing in the GSM8K test set.}
    \label{tab:decontamination_results}
\end{table*}

\vspace{-0.5em}
\paragraph{GSM-Symbolic.}
Table~\ref{tab:gsm_symbolic_results} shows the performance of the instruction-tuned models on GSM-Symbolic. The performance of these models on the original problems is presented in the GSM8k column in the table. We observe that the model fine-tuned on the \tulu{} mixture with the arithmetic dataset sees a performance drop of 25.9\% compared to a 44.4\% drop for the model fine-tuned only on the \tulu{} mixture, indicating greater robustness to numerical changes in the original problems. We observe a similar pattern with greedy decoding. See Appendix~\ref{app:robustness_greedy} for details. These results indicate that including arithmetic examples in the fine-tuning mixture makes models more robust to numerical perturbations.

\section{Discussion}
\subsection{Impact on General Abilities}
In this section, we investigate the impact of incorporating the arithmetic examples in the training pipeline, given the models' limited capacity. Because targeted fine-tuning has been shown to cause models to forget previously learned abilities, we exclude models from Section~\ref{sec:intermediate_fine_tuning}, which are directly fine-tuned on \gsm{}, and focus solely on instruction-tuned models in this analysis. We evaluate the instruction-tuned models from Section~\ref{sec:instruction_tuning} on IFEval \citep{zhou2023instruction} and BigBench-Hard \citep{suzgun2022challenging}. The performance of these models is shown in Table~\ref{tab:ifeval_bbh_results}. The IFEval results show that the impact on instruction-following abilities is minimal. On the other hand, the BigBench-Hard results indicate that model performance on other reasoning tasks is adversely impacted by including arithmetic examples in the SFT mixture.

\subsection{Matching the Token Budget}
The synthetic arithmetic dataset is one-tenth of the \tulu{} mixture in terms of the number of tokens. For this ablation study, we remove 10\% of the randomly selected examples from the \tulu{} mixture to match the token budget of the instruction-tuning run without the arithmetic dataset. We fine-tune GPT2-Large on a mixture of this dataset and the arithmetic dataset using the same pipeline as used in Section~\ref{sec:instruction_tuning}. Table~\ref{tab:token_budget_results} shows the performance of the resulting model. The performance on \gsm{} and GSM-Plus degrades compared to the models instruction-tuned on all examples from the \tulu{} mixture (Table~\ref{tab:inst_tuning_results}). These findings suggest that while the synthetic arithmetic dataset can help the model learn arithmetic computations, it cannot replace examples that help models develop broader reasoning abilities.

\subsection{Data Contamination Analysis}
Since the arithmetic dataset is generated programmatically, some of its arithmetic computations may overlap with those found in the downstream mathematical reasoning datasets. To ensure that the observed performance improvements are not simply due to such overlap, we investigate the extent to which potential data contamination may influence the results.

To eliminate dataset contamination as a possible reason for the improved performance in our experiments, we remove from the synthetic arithmetic dataset examples that contain arithmetic computations appearing in the GSM8k test set. The only exception to this is the computations involving single-digit numbers, as these are fundamental operations that the model is expected to memorize. We repeat the intermediate fine-tuning experiments described in Section~\ref{sec:intermediate_fine_tuning}, where we fine-tune FlanT5-Base and GPT2 on this dataset before subsequently fine-tuning them on \gsmd{}. The results are shown in Table~\ref{tab:decontamination_results}, which are largely similar to those reported in Table~\ref{tab:main_results}. These results indicate that the models do not merely memorize arithmetic computations, allowing us to rule out data contamination as the cause of the performance improvements observed in our experiments.

\section{Conclusion}
In this work, we explored the impact of incorporating an arithmetic dataset through two approaches: intermediate fine-tuning and integration within the instruction-tuning mixture. Our experiments demonstrated that both approaches boost mathematical reasoning performance in smaller models. While intermediate fine-tuning can make subsequent fine-tuning on other tasks more challenging, the issue can be mitigated by using a larger dataset. Additionally, we found that models fine-tuned on the arithmetic-inclusive mixture demonstrate a clear performance advantage over those trained on a standard instruction set. These findings highlight the crucial role of explicit arithmetic training in strengthening mathematical reasoning in smaller models.

\section{Limitations}
\label{sec:limitations}
While incorporating an arithmetic dataset improves a model's mathematical reasoning performance, smaller models still have considerable room for improvement. In this work, we include the arithmetic dataset in the training pipeline but do not investigate other factors, such as custom embedding schemes for arithmetic computation. A promising direction for future research is to incorporate insights from recent work, such as \citet{mcleish2024transformers}, into the model architecture. Another limitation pertains to the instruction-tuning mixture. While including the arithmetic dataset improves few-shot performance on mathematical reasoning benchmarks, data mixture ablations may be necessary to optimize the instruction-tuning mixture for better overall performance. Finally, although this study focuses on smaller models, its findings apply to larger models. The arithmetic capabilities of pre-trained models could be further enhanced by leveraging synthetic arithmetic datasets. We leave these explorations for future work.

\section{Acknowledgments}
This work was supported by the Illinois Campus Cluster Program and the Delta Advanced Computing and Data Resource. Delta is supported by the National Science Foundation (award OAC 2005572) and the State of Illinois and is a joint effort of the University of Illinois Urbana-Champaign and its National Center for Supercomputing Applications.
\nocite{access}

\section{Bibliographical References}\label{sec:reference}
\bibliographystyle{lrec2026-natbib}
\bibliography{custom}

\appendix
\clearpage

\section{Synthetic Arithmetic Dataset}
\label{app:synthetic_arithmetic_dataset}
The dataset consists of problems of addition, subtraction, multiplication, division, fractions, and percentages, all with two operands. These problems do not require any reasoning and are simple arithmetic problems. Table~\ref{tab:arith_dataset_samples} shows examples from each category.

\section{Training Details}
\subsection{Intermediate Fine-Tuning}
\label{app:training_details_ift}
We use the AdamW optimizer \citep{loshchilov2017decoupled} with a learning rate of $10^{-4}$, a weight decay of $10^{-4}$, and an effective batch size of 128. For FlanT5-Large and GPT2-Large, a learning rate warmup of 500 steps is used. The intermediate fine-tuning is performed for two epochs without validation. To adapt these models for reasoning, we continue the training from these checkpoints on \gsm{}. The models are fine-tuned for 20 and 100 epochs on \gsmd{} and \gsmo{}, respectively. The best checkpoint is selected based on the \gsm{} validation performance.

\subsection{Instruction Tuning}
\label{app:training_details_it}
We use the AdamW optimizer \citep{loshchilov2017decoupled} with $2 \times 10^{-4}$ learning rate and a weight decay of $10^{-4}$. A learning rate warmup of 500 steps is used. As examples in the \tulu{} mixture have varied sequence lengths and differ significantly from the arithmetic examples in this regard, we use a variable batch size with approximately 0.5M tokens in each batch. The input to the model is truncated from the left to have at most 1024 tokens.

\section{Arithmetic in Reasoning Context}
\label{app:gsm8k_arith_details}
We evaluate the frequency of numerical computation errors made by models within a reasoning context. For this evaluation, we use the \gsm{} test set and identify arithmetic computations based on the provided calculation annotations (enclosed within \texttt{<<>>}). Figure~\ref{fig:gsm_arith_examples} shows examples used for this evaluation. Given a question and its solution up to an annotation, models are prompted to generate five tokens. We extract the numerical value at the beginning of the generated text and compare it to the expected output to measure accuracy.
\begin{table}[t]
    \centering
    \begin{small}
    \begin{tabular}{ll}
    \toprule
    Type & Problem \\
    \midrule
    Addition & Find the value of 3 + 1872. \\
    Subtraction & Solve 90352 - 19621. \\
    Multiplication & Compute the value of 552 x 4. \\
    Division & Determine 832 / 2. \\
    Fractions & 1 / 19 of 573097= \\
    Percentages & Calculate 1\% of 3200 \\
    \bottomrule
    \end{tabular}
    \end{small}
    \caption{Samples from the synthetic arithmetic dataset.}
    \label{tab:arith_dataset_samples}
\end{table}
\begin{figure}[t]
    \centering
    \scalebox{0.9}{
    \begin{tikzpicture}[
    font=\footnotesize\fontfamily{phv}\selectfont,
    text=darkgray!90,
]
    \node [fill=CadetBlue!15, rounded corners, text width=\linewidth] (ORIG_EX) {
    {\bfseries Question:} Dylan needed chicken sausages and fish sausages to make sausage buns at a party. He bought 38 chicken sausages and 6 more fish sausages than chicken sausages. How many sausages did Dylan buy in all?\\

    {\bfseries Answer:} He bought 38 + 6 = <<38+6=44>>44 fish sausages. Dylan bought 38 + 44 = <<38+44=82>>82 sausages in all. \#\#\#\# 82
    };

    \node [anchor=north west] at ($(ORIG_EX.north west) + (0, 15pt)$) {\bfseries Original Example:};

    \node [anchor=north west] at ($(ORIG_EX.north west) + (0, -105pt)$) {\bfseries Examples for GSM8k Arithmetic Evaluation:};

    \node [fill=OliveGreen!10, rounded corners, below=of ORIG_EX, text width=\linewidth] (Q1) {
    {\bfseries Model Input:}\\
    Question: Dylan needed chicken sausages and fish sausages to make sausage buns at a party. He bought 38 chicken sausages and 6 more fish sausages than chicken sausages. How many sausages did Dylan buy in all?\\
    Let's think step by step\\
    He bought 38 + 6 =\\
    \vspace{0.75em}
    {\bfseries Expected Output:}\\
    44
    };

    \node [fill=BurntOrange!15, rounded corners, below=of Q1, text width=\linewidth, below=0.2cm] (Q2) {
    {\bfseries Model Input:}\\
    Question: Dylan needed chicken sausages and fish sausages to make sausage buns at a party. He bought 38 chicken sausages and 6 more fish sausages than chicken sausages. How many sausages did Dylan buy in all?\\
    Let's think step by step\\
    He bought 38 + 6 = 44 fish sausages. Dylan bought 38 + 44 =\\
    \vspace{0.75em}
    {\bfseries Expected Output:}\\
    82
    };
\end{tikzpicture}
    }
    \caption{An example of how models are evaluated for arithmetic errors in reasoning contexts. This example has two arithmetic computations, each resulting in a test example for the GSM8k arithmetic evaluation.}
    \label{fig:gsm_arith_examples}
\end{figure}
\begin{figure*}[p]
\begin{lstlisting}[
        breaklines,
        basicstyle=\scriptsize\ttfamily,
        columns=flexible,
        captionpos=b,
        caption=Prompt for math word problem datasets except for AQuA.,
        backgroundcolor=\color{Gray!10},
        frame=tlbr,
        framesep=5pt,
        % numbers=left,
        label=fig:mwp_prompt,
    ]
Question: There are 15 trees in the grove. Grove workers will plant trees in the grove today. After they are done, there will be 21 trees. How many trees did the grove workers plant today?
Let's think step by step
There are 15 trees originally.
Then there were 21 trees after some more were planted.
So there must have been 21 - 15 = 6.
The answer is 6.

Question: If there are 3 cars in the parking lot and 2 more cars arrive, how many cars are in the parking lot?
Let's think step by step
There are originally 3 cars.
2 more cars arrive.
3 + 2 = 5.
The answer is 5.

Question: Leah had 32 chocolates and her sister had 42. If they ate 35, how many pieces do they have left in total?
Let's think step by step
Originally, Leah had 32 chocolates.
Her sister had 42.
So in total they had 32 + 42 = 74.
After eating 35, they had 74 - 35 = 39.
The answer is 39.

Question: Jason had 20 lollipops. He gave Denny some lollipops. Now Jason has 12 lollipops. How many lollipops did Jason give to Denny?
Let's think step by step
Jason started with 20 lollipops.
Then he had 12 after giving some to Denny.
So he gave Denny 20 - 12 = 8.
The answer is 8.
\end{lstlisting}
\end{figure*}
\begin{figure*}[p]
\begin{lstlisting}[
        breaklines,
        basicstyle=\scriptsize\ttfamily,
        columns=flexible,
        captionpos=b,
        caption=Prompt for AQuA.,
        backgroundcolor=\color{Gray!10},
        frame=tlbr,
        framesep=5pt,
        % numbers=left,
        label=fig:aqua_prompt,
    ]
Q: John found that the average of 15 numbers is 40. If 10 is added to each number then the mean of the numbers is?
Answer Choices: (a) 50 (b) 45 (c) 65 (d) 78 (e) 64
A: If 10 is added to each number, then the mean of the numbers also increases by 10. So the new mean would be 50. The answer is (a).

Q: If a / b = 3/4 and 8a + 5b = 22,then find the value of a.
Answer Choices: (a) 1/2 (b) 3/2 (c) 5/2 (d) 4/2 (e) 7/2
A: If a / b = 3/4, then b = 4a / 3. So 8a + 5(4a / 3) = 22. This simplifies to 8a + 20a / 3 = 22, which means 44a / 3 = 22. So a is equal to 3/2. The answer is (b).

Q: A person is traveling at 20 km/hr and reached his destiny in 2.5 hr then find the distance?
Answer Choices: (a) 53 km (b) 55 km (c) 52 km (d) 60 km (e) 50 km
A: The distance that the person traveled would have been 20 km/hr * 2.5 hrs = 50 km. The answer is (e).

Q: How many keystrokes are needed to type the numbers from 1 to 500?
Answer Choices: (a) 1156 (b) 1392 (c) 1480 (d) 1562 (e) 1788
A: There are 9 one-digit numbers from 1 to 9. There are 90 two-digit numbers from 10 to 99. There are 401 three-digit numbers from 100 to 500. 9 + 90(2) + 401(3) = 1392. The answer is (b).
\end{lstlisting}
\end{figure*}
\begin{table*}[p]
    \centering
    \begin{small}
    \begin{tabular}{ccrrrrr}
    \toprule
    \textbf{\tulu{}} & \textbf{Arith.} & \textbf{MultiArith} & \textbf{AddSub} & \textbf{SingleOp} & \textbf{SingleEq} & \textbf{SimulEq} \\
    \midrule
    \multicolumn{7}{c}{\it Greedy Decoding} \\
    \midrule
    \xmark & \xmark & 2.8 & 2.8 & 4.4 & 1.8 & 1.4 \\
    \cmark & \xmark & 39.4 & 11.9 & 45.3 & 43.1 & \textbf{6.8} \\
    \cmark & \cmark & \textbf{50.0} & \textbf{36.7} & \textbf{64.2} & \textbf{59.6} & 4.8 \\
    \midrule
    \multicolumn{7}{c}{\it Self-Consistency Decoding} \\
    \midrule
    \xmark & \xmark & 3.7 & 1.2 & 2.7 & 0.6 & 2.5 \\
    \cmark & \xmark & 56.1 & 18.7 & 51.4 & 47.7 & 5.3 \\
    \cmark & \cmark & \textbf{65.9} & \textbf{41.3} & \textbf{68.3} & \textbf{65.7} & \textbf{5.5} \\
    \bottomrule
    \end{tabular}
    \end{small}
    \caption{Accuracy (\%) achieved by the instruction-tuned GPT2-Large models on datasets in MAWPS. The first rows under both greedy and self-consistency decoding denote the pre-trained model.}
    \label{tab:mawps_detailed_results}
\end{table*}

\section{Prompt for Math Word Problems}
\label{app:math_prompt}
We use four exemplars from the prompt from \citet{wei2022chain} to evaluate the models on math word problem datasets except AQuA. The prompt is shown in Listing~\ref{fig:mwp_prompt}. The prompt for AQuA is shown in Listing~\ref{fig:aqua_prompt}.

\section{MAWPS Detailed Results}
\label{app:mawps_detailed_results}
Table~\ref{tab:mawps_detailed_results} shows the accuracy (\%) achieved by the instruction-tuned models on different datasets within MAWPS.

\section{Robustness with Greedy Decoding}
\label{app:robustness_greedy}
We also evaluate the robustness of post-trained models with greedy decoding. Our results show that the model fine-tuned on the \tulu{} mixture and the arithmetic dataset is more robust to numerical changes than the model fine-tuned only on the \tulu{} mixture. The results on GSM-Plus and GSM-Symbolic are illustrated in Figure~\ref{fig:gsm_plus_accuracy_greedy} and Table~\ref{tab:gsm_symbolic_results_greedy}, respectively.

\begin{figure}[]
    \centering
    \resizebox{\linewidth}{!}{
        \large\fontfamily{phv}\selectfont
        \includesvg[]{img/gsm_plus_category_wise_greedy.svg}
    }
    \caption{Performance of the pre-trained and instruction-tuned GPT2-Large models on GSM-Plus for different perturbation types using greedy decoding. The model fine-tuned on \tulu{} mixture and the arithmetic dataset performs better across different perturbation types. The percentages above the bars represent the performance drop relative to the original GSM8k dataset, as shown in Table~\ref{tab:inst_tuning_results}.}
    \label{fig:gsm_plus_accuracy_greedy}
\end{figure}

\begin{table}[]
    \centering
    \begin{small}
    \begin{tabular}{ccrrr}
    \toprule
    \tulun{} & \multirow{2}{*}{Arith.} & GSM8k & GSM-Symb. & $\Delta$ \\
    SFT & {} & Acc. (\%) & Acc. (\%) & (\%) \\
    \midrule
    \cmark & \xmark & 10.0 & 6.8 $\pm$ 1.9 & $-$32.0 \\
    \cmark & \cmark & 17.0 & 13.3 $\pm$ 2.6 & \textbf{$-$21.8} \\
    \bottomrule
    \end{tabular}
    \end{small}
    \caption{Performance of the post-trained models using greedy decoding when evaluated on the original GSM8k problems vs the same problems in GSM-Symbolic. The performance drop is indicated by $\Delta$.}
    \label{tab:gsm_symbolic_results_greedy}
\end{table}

\begin{table*}[]
    \centering
    \setlength{\tabcolsep}{2pt}
    \begin{small}
    \begin{tabular}{lrr}
    \toprule
    \textbf{Task} & \textbf{w/o Arit.} & \textbf{w/ Arith.} \\
    \midrule
    \texttt{boolean\_expressions} & 55.20 & 53.20 \\
    \texttt{causal\_judgement} & 13.37 & 10.70 \\
    \texttt{date\_understanding} & 17.20 & 16.80 \\
    \texttt{logical\_deduction\_five\_objects} & 15.20 & 16.80 \\
    \texttt{logical\_deduction\_seven\_objects} & 5.60 & 0.40 \\
    \texttt{logical\_deduction\_three\_objects} & 27.20 & 28.80 \\
    \texttt{movie\_recommendation} & 8.00 & 16.40 \\
    \texttt{navigate} & 58.40 & 42.40 \\
    \texttt{object\_counting} & 8.40 & 2.00 \\
    \texttt{penguins\_in\_a\_table} & 10.27 & 10.96 \\
    \texttt{reasoning\_about\_colored\_objects} & 6.40 & 12.00 \\
    \texttt{snarks} & 13.48 & 20.22 \\
    \texttt{sports\_understanding} & 55.60 & 47.60 \\
    \texttt{tracking\_shuffled\_objects\_five\_objects} & 11.60 & 10.80 \\
    \texttt{tracking\_shuffled\_objects\_seven\_objects} & 0.40 & 0.40 \\
    \texttt{tracking\_shuffled\_objects\_three\_objects} & 30.00 & 33.20 \\
    \texttt{web\_of\_lies} & 4.80 & 1.60 \\
    \texttt{word\_sorting} & 1.60 & 0.40 \\
    \bottomrule
    \end{tabular}
    \end{small}
    \caption{Detailed BigBench-Hard performance of GPT2-Large instruction-tuned on the \tulu{} mixture, with and without the inclusion of the arithmetic dataset.}
    \label{tab:bbh_detailed_results}
\end{table*}

\section{Dataset Statistics}
Table~\ref{tab:dataset_stats} shows the statistics of the datasets used in this work.

\begin{table*}[]
    \centering
    \setlength{\tabcolsep}{3pt}
    \begin{small}
    \begin{tabular}{lrll}
    \toprule
    Dataset & \# Examples & \multicolumn{2}{l}{Source} \\
    \midrule
    \multicolumn{4}{l}{\it Training and Validation} \\
    \midrule
    \gsmo{} & 6,961 & \textsc{hf} & \texttt{openai/gsm8k} \\
    \gsmd{} & 175,668 & - & {} \\
    Arithmetic & 1,290,175 & - & {} \\
    \tulu{} Mix. & 896,090 & \textsc{hf} & \texttt{allenai/tulu-3-sft-mixture} \\
    \gsm{} (val.) & 512 & \textsc{hf} & \texttt{openai/gsm8k} \\
    \midrule
    \multicolumn{4}{l}{\it Test} \\
    \midrule
    \gsm{} & 1,319 & \textsc{hf} & \texttt{openai/gsm8k} \\
    ASDiv & 2,305 & \textsc{gh} & \texttt{chaochun/nlu-asdiv-dataset} \\
    SVAMP & 1,000 & \textsc{gh} & \texttt{arkilpatel/SVAMP} \\
    MultiArith & 180 & \textsc{hf} & \texttt{ChilleD/MultiArith} \\
    AddSub & 109 & \textsc{hf} & \texttt{allenai/lila} \\
    SingleOp & 159 & \textsc{hf} & \texttt{allenai/lila} \\
    SingleEq & 109 & \textsc{hf} & \texttt{allenai/lila} \\
    SimulEq & 146 & \textsc{hf} & \texttt{allenai/lila} \\
    AQuA & 254 & \textsc{gh} & \texttt{google-deepmind/AQuA} \\
    GSM-Plus & 10,552 & \textsc{hf} & \texttt{qintongli/GSM-Plus} \\
    GSM-Symbolic & 100 $\times$ 50 & \textsc{gh} & \texttt{apple/ml-gsm-symbolic} \\
    \bottomrule
    \end{tabular}
    \end{small}
    \caption{The number of examples in the datasets used in this work and their sources. \textsc{hf} and \textsc{gh} indicate HuggingFace and GitHub, respectively.}
    \label{tab:dataset_stats}
\end{table*}

\end{document}